\documentclass[11pt,a4paper]{article}
\usepackage[hyperref]{acl2021}
\usepackage{times}
\usepackage{latexsym}

\usepackage[T1]{fontenc}
\usepackage[utf8]{inputenc}
\usepackage{microtype}
\usepackage{multirow}
\usepackage{booktabs}
\usepackage{threeparttable}
\usepackage{graphicx}
\usepackage{subcaption}
\usepackage{mathtools}
\usepackage{algorithm}
\usepackage{algpseudocode}
\usepackage{adjustbox}
\usepackage{amssymb}

\aclfinalcopy 
\def\aclpaperid{2673}

\newcommand{\tmix}{TMix}
\newcommand{\emix}{EmbedMix}

\newcommand{\ours}{\textit{SSMix}} 

\newcommand{\xzero}[1][]{x^{A}_{#1}}
\newcommand{\xone}[1][]{x^{B}_{#1}}
\newcommand{\yzero}{y^{A}} 
\newcommand{\yone}{y^{B}} 
\newcommand{\batchone}{A} 
\newcommand{\batchtwo}{B}

\title{
\ours{}: {S}aliency-{B}ased {S}pan {M}ixup for {T}ext {C}lassification
}

\author{
    Soyoung Yoon$^{1,2}$\thanks{\, Equal contribution.}\,\,\thanks{\, Work done during the internship at Clova AI.} \quad
    Gyuwan Kim$^1$\footnotemark[1] \quad 
    Kyumin Park$^2$ \\
    $^1$Clova AI, Naver Corp. \quad $^2$KAIST \\
    \texttt{\{soyoungyoon,pkm9403\}@kaist.ac.kr, gyuwan.kim@navercorp.com} \\
}

\date{}

\begin{document}
\maketitle
\begin{abstract}
Data augmentation with mixup has shown to be effective on various computer vision tasks. 
Despite its great success, there has been a hurdle to apply mixup to NLP tasks since text consists of discrete tokens with variable length. 
In this work, we propose \ours{}, a novel mixup method where the operation is performed on input text rather than on hidden vectors like previous approaches.
\ours{} synthesizes a sentence while preserving the locality of two original texts by span-based mixing and keeping more tokens related to the prediction relying on saliency information.
With extensive experiments, we empirically validate that our method outperforms hidden-level mixup methods on a wide range of text classification benchmarks, including textual entailment, sentiment classification, and question-type classification. 
Our code is available at \href{https://github.com/clovaai/ssmix}{https://github.com/clovaai/ssmix}.
\end{abstract}

\section{Introduction}

Data augmentation gains popularity in natural language processing (NLP) \citep{feng2021survey} due to the expensive cost of data collection.
Some of them are based on simple rules \cite{wei2019eda} and models \cite{edunov2018understanding, ng2020ssmba} to generate similar text.
Augmented samples are trained jointly with original samples by a standard way or advanced training methods \cite{zhu2019freelb, park2021consistency}.
On the other hand, mixup \citep{zhang2018mixup} interpolates input texts and labels for the augmentation.

Training with mixup and its variants become a popular regularization method in computer vision to improve the generalization of neural networks. 
% due to their effectiveness to improve the accuracy.
Mixup approaches are categorized into input-level mixup \citep{yun2019cutmix, kim2020puzzle, walawalkar2020attentive, uddin2021saliencymix} and hidden-level mixup \citep{verma2019manifold} depending on the location of the mix operation.  % generate virtual examples by mixing two sampled inputs along with their labels
Input-level mixup is a more prevalent approach than hidden-level mixup because of its simplicity and the ability to capture locality, leading to better accuracy.

Applying mixup in NLP is more challenging than in computer vision because of the discrete nature of text data and variable sequence lengths.
Therefore, most previous attempts on mixup for texts \citep{guo2019augmenting, chen-etal-2020-mixtext} apply mixup on hidden vectors like embeddings or intermediate representations. % because mixing two discrete token sequences with variable lengths are non-trivial.
However, input-level mixup might have an advantage over hidden-level mixup with a similar intuition from computer vision.
This motivation encourages us to examine input-level mixup approaches for text data.

\begin{figure}[t!]
    \centering
    \begin{adjustbox}{max size={0.45\textwidth}{0.45\textheight}}
    \includegraphics{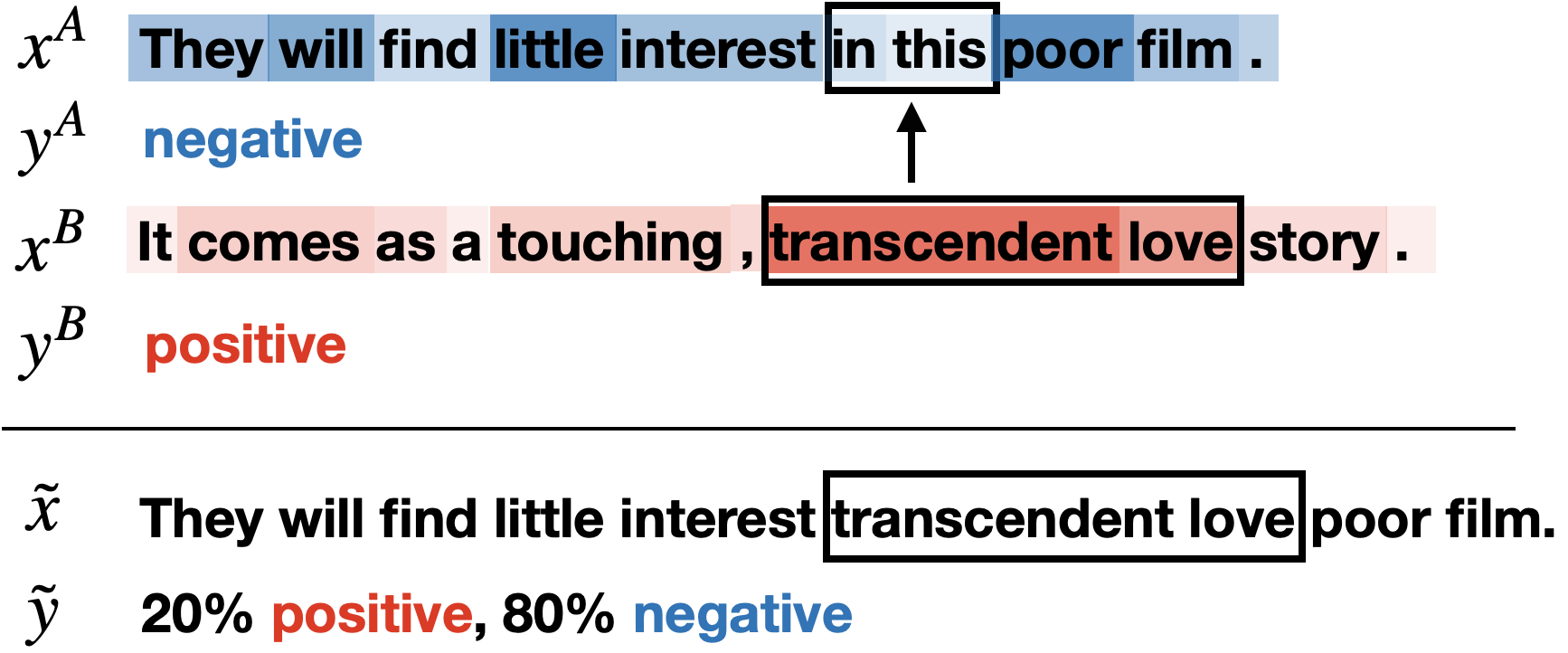}
    \end{adjustbox}
    \caption{
    Illustration of \ours{}. 
    Two data samples $\xzero{}$ and $\xone{}$ are labeled negative and positive respectively for sentiment classification task. For each token, saliency maps are visualized where darker concentration of colors mean higher contribution to corresponding label. We select the least salient span from $\xzero{}$ and replace it with the most salient span from $\xone{}$. The output results in $\tilde{x} = mixup(\xzero{}, \xone{})$. We also assign $\tilde{y}$ by the mixup ratio $\lambda$. In this example, $\lambda$ is set to 0.2 as the span length is 2 out of 10. }
    \label{fig:salmix}
    \vspace{-15pt}
\end{figure}

In this work, we propose \ours{} (Fig~\ref{fig:salmix}), a novel input-level spanwise mixup method considering the saliency of spans. 
% \ours{} contains two characteristics. 
First, we conduct a mixup by replacing a span of contiguous tokens with a span in another text, which is inspired from CutMix \citep{yun2019cutmix}, to preserves the locality of two source texts in the mixed text. 
% Another point is span selection from saliency. 
Second, we select a span to be replaced and to replace based on saliency information to make the mixed text contain tokens more related to output prediction, which may be semantically important.
% the least salient span among one text as a replaced data, and the most salient span in another text as a replacing data.
% Our mixup method is illustrated in Fig~\ref{fig:salmix}.
% Aforementioned characteristics lead to improvement in mixup performance in several perspectives. 
% From the replacement between spans, our method preserves the locality of two source texts in mixed text. 
% Also, saliency-based span selection generates mixed sentence containing tokens more related to output prediction, which may be semantically important.
Our input-level method is different from hidden level mixup methods in that while current hidden level mixup methods linear interpolate original hidden vectors, our method mix tokens on the input level, resulting in a \textit{nonlinear} output. Also, we utilize saliency values to select span from each sentence and discretely define the length of span and mixup ratio, which is outside the hidden level.

\ours{} has empirically proven effective through extensive experiments on a wide range of text classification benchmarks. 
Especially, we prove that input-level mixup methods generally outperform hidden-level methods. 
We also show the importance of using saliency information and restricting token selection in span-level when conducting our method via ablation study. 

\section{\ours{}}
\label{section:salmix}
We propose \ours{} to synthesize a new text $\tilde{x}$ by replacing a span $\xzero[S]$ from one text $\xzero$ into another span $\xone[S]$ from another text $\xone$ based on saliency information.
Also, we have to set a new label $\tilde{y}$ for $\tilde{x}$ using $\yzero$ and $\yone$ which are one-hot labels corresponding to $\xzero$ and $\xone$, respectively.
Consequently, we can additionally use this generated virtual sample $(\tilde{x}, \tilde{y})$ for training.  

\paragraph{Saliency}
Saliency measures how each portion of data (in this case, tokens) affects the final prediction.
Gradient-based methods \citep{simonyan2013deep, li-etal-2016-visualizing} are widely used for the saliency computation.
We compute the gradient of classification loss $\mathcal{L}$ with respect to input embedding $e$, and use its magnitude as the saliency: i.e., $s = \left\|\partial \mathcal{L} / \partial e \right\|_{2}$. 
We apply the L2 norm to obtain the magnitude of a gradient vector, which becomes a saliency of each token similar to PuzzleMix \cite{kim2020puzzle}.

\paragraph{Mixing text}
Text data $\xzero$ and $\xone$ are discrete token sequences.
Using saliency scores as explained earlier, we can find the least salient span in $\xzero$ with a length $l_A$ as $\xzero[S]$ and the most salient span in $\xzero$ with a length $l_B$ as $\xone[S]$.
We set $l_A = l_B = max(min([\lambda_{0}|\xzero|], |\xone|), 1)$ given a prior mixup  ratio $\lambda_{0}$.
Then, final $\tilde{x}$ becomes the concatenation of $(\xzero[L]; \xone[S]; \xzero[R])$ where $\xzero[L]$ and $\xzero[R]$ are tokens located to the left and the right side of $\xzero[S]$ respectively in the original text $\xzero$.
\begin{algorithmic}
\begin{algorithm}[!t]
\caption{Mixup loss calculation}
\Procedure{SSMix\_Loss}{$\xzero{}, \xone{}, \yzero{}, \yone{}, \lambda$}
   \State $\tilde{x} \gets SSMix(\xzero{}, \xone{})$
   \State $logit \gets model(\tilde{x})$
   \State $loss^{A} \gets CrossEntropy(logit, y^{A})$ 
   \State $loss^{B} \gets CrossEntropy(logit, y^{B})$
   \State $total\_loss \gets loss^{A} * \lambda + loss^{B} * (1-\lambda)$
   \Return $total\_loss$
\EndProcedure
\label{alg:loss_calculation}
\end{algorithm}
\end{algorithmic}

\paragraph{Same span length}
We set the length of the original ($l_A$) and replaced ($l_B$) span to be the same, since allowing different length of spans would result in redundant and ambiguous mixup variations. Also, calculating the mixup ratio between different span length would be too complex. This same-size replacement strategy is also adopted in \citet{yun2019cutmix} and \citet{uddin2021saliencymix}. In situations where span length is the same, our method maximizes the effect of saliency. Since \ours{} doesn't restrict the position of tokens, we can pick the \textit{most} salient span and replace it with \textit{least} salient span on the other text.

\paragraph{Mixing label}
We set mixup ratio $\lambda$ for label as $\lambda = |\xone[S]| / |\tilde{x}|$. 
Since $\lambda$ is recalculated by counting the number of tokens in the span, it may differ from $\lambda_{0}$.
We set the label of $\tilde{x}$ to $\tilde{y} = (1 - \lambda) \yzero + \lambda \yone$.
%Upon implementation, we calculate the cross-entropy loss with respect to each label from original pairs and do a weighted sum to calculate the loss for SSMix. Algorithm \ref{alg:loss_calculation} shows the pseudo-code for SSMix in detail.
Algorithm~\ref{alg:loss_calculation} shows how we utilize the original sample pairs to compute the mixup loss for augmented samples.
We calculate the cross-entropy loss of the augmented output logit with respect to the original target label of each sample and combine them by weighted sum, which is similar to the original implementation of \citet{zhang2018mixup}.\footnote{\href{https://github.com/hongyi-zhang/mixup/blob/master/cifar/utils.py\#L34}{https://github.com/hongyi-zhang/mixup/blob/master/cifar/utils.py\#L34}}
Therefore, applying \ours{} is independent of the total number of labels of the classification dataset. On any dataset, output label ratio is calculated by linear combination of \textit{two} original labels.

\paragraph{Paired sentence tasks}
For tasks requiring a pair of texts as an input such as textual entailment and similarity classification,
we conduct mixup in a pairwise manner and calculate the mixup ratio by aggregating token counts in each mixup result. 
Denoting $\xzero = (p^{A}, q^{A})$, $\xone = (p^{B}, q^{B})$, and $\tilde{x} = (\tilde{p}, \tilde{q})$, we define mixup of paired sentence data as $\tilde{x} = (mixup(p^{A}, p^{B}), mixup(q^{A}, q^{B}))$. Here, we set the mixup ratio on paired sentence tasks as $\lambda = (|p_{S}| + |q_{S}|) / (|\tilde{p}| + |\tilde{q}|)$, where $p_{S}$ and $q_{S}$ are replacing spans of independent mixup operations. Illustration is available in Appendix \ref{appendix:paired_example}.

\section{Experimental Setup}

\subsection{Dataset}
\begin{table}[!t]
    \centering
    \begin{threeparttable}
        \resizebox{0.48\textwidth}{!}{
        \begin{tabular}{lccc}
            \toprule
            \multicolumn{1}{c}{Dataset} & Task & $\#$ Label & Size  \\
            \midrule
            SST-2 & Sentiment & 2 & 67k / 1.8k \\
            \midrule
            QQP & Paraphrase & 2 & 364k / 391k \\
            \midrule
            MNLI & NLI & 3 & 393k / 20k \\
            \midrule
            QNLI & QA/NLI & 3 & 105k / 5.4k  \\
            \midrule
            RTE & NLI & 2 & 2.5k / 3k  \\
            \midrule
            MRPC & Paraphrase & 2 & 3.7k / 1.7k  \\
            \midrule
            TREC-coarse & Classification & 6 & 5.5k / 500  \\
            \midrule
            TREC-fine & Classification & 47 & 5.5k / 500  \\
            \midrule
            ANLI & NLI & 3 & 162.8k / 3.2k / 3.2k  \\
            \bottomrule
        \end{tabular}}
    \end{threeparttable}
    \caption{Dataset name, task, number of total labels, and dataset size of datasets we used as benchmark. Task column describes the objective of each dataset. ANLI dataset shows aggregated dataset statistics among different rounds. GLUE tasks report the size as (train / validation) format, TREC reports (train / test) and ANLI reports (train / validation / test).}
    \label{tab:dataset_summary}
\end{table}
As listed in table \ref{tab:dataset_summary}, to evaluate the effectiveness of \ours{}, we perform experiments on various text classfication benchmarks: six datasets in GLUE benchmark \citep{wang-etal-2018-glue}, TREC \citep{li-roth-2002-learning, hovy-etal-2001-toward}, and ANLI \cite{nie2019adversarial}.
Two of them are single sentence classification tasks, and six of them are sentence pair classification tasks.
All datasets are extracted from HuggingFace datasets library.\footnote{\href{https://github.com/huggingface/datasets}{https://github.com/huggingface/datasets}}

For GLUE, we use
SST-2 \citep{socher-etal-2013-recursive}, MNLI \citep{williams-etal-2018-broad}, QNLI \citep{DBLP:journals/corr/RajpurkarZLL16}, RTE \citep{bentivogli2009fifth}, MRPC \citep{dolan-brockett-2005-automatically}, and QQP\footnote{\href{https://www.quora.com/First-Quora-Dataset-Release-Question-Pairs}{https://www.quora.com/First-Quora-Dataset-Release-Question-Pairs}}.
Among GLUE, we leave out datasets that were not evaluated by accuracy, along with WNLI, because the size is too small to show any general trend of effectiveness.

TREC is a commonly used dataset to evaluate mixup methods in sentence classification \cite{guo2019augmenting, thulasidasan2019mixup}.
We use two different versions of TREC (coarse, fine) that have different levels of label number to test the dependency of mixup effectiveness on the number of class labels. 
In addition, we use ANLI to see how mixup can help to improve model robustness. 
For training ANLI, we concatenate all training data from different rounds and use them to train the model.

\subsection{Baseline}
We compare \ours{} with three baselines: (1) standard training without mixup, (2) \emix{}, and (3) \tmix{}.
\emix{} apply mixup on the embedding layer, which is similar to the wordMixup in \citet{guo2019augmenting} except their experiments are performed with LSTM or CNN architecture. 
\tmix{}, borrowed from \citet{chen-etal-2020-mixtext}, interpolates hidden states of two different inputs at a particular encoder layer and forward the combined hidden states to the remaining layers.
For \emix{} and \tmix{}, we follow the best settings stated in the original papers: mixup ratio is set by $\lambda' \sim Beta(\alpha, \alpha)$, $\lambda = max(\lambda', 1-\lambda')$ with $\alpha=0.2$.
During the training with \tmix{}, we randomly sample the mixup layer from $[7, 9, 12]$. 

\subsection{Ablation study}
To investigate how much (1) considering saliency and (2) restricting mixup operation on the span-level individually benefit our proposed method, we conduct an ablation study. 
We implement \ours{} without considering saliency information (\ours{} - saliency) where the spans are randomly selected, and additionally without the span-level restriction (\ours{} - saliency - span). For \ours{} - saliency - span, we randomly sample tokens from $\xone{}$, which need not be a contiguous span and are conducted on a per-token basis. Then, we replace tokens accordingly with the position of the token be preserved, meaning that the second token from $\xzero{}$ is replaced with the second token from $\xone{}$, and so on.
For all ablation studies, the lambda values were set to 0.1 to compare methods with the same setting as \ours{}. 
Detailed implementation and illustration of ablation methods and comparison with simple word dropout methods are described in Appendix~\ref{section:ablation_general}.

\subsection{Training Details}
Among the entire experiment, we use sequence classification task with the pre-trained BERT-base model having 110M parameters from HuggingFace Transformers library.\footnote{\href{https://github.com/huggingface/transformers}{https://github.com/huggingface/transformers}}
We perform all experiments with five different seeds (0 to 4) on a single NVIDIA P40 GPU and report the average score.
We set a maximum sequence length of 128, batch size of 32, with AdamW optimizer with eps of 1e-8 and weight decay of 1e-4. 
We use a linear scheduler with a warmup for 10\% of the total training step. We update the best checkpoint by measuring validation accuracy on every 500 steps. For datasets that have less than 500 steps per epoch, we update and validate every epoch.

Considering our objective of enhancing performance through mixup, we conduct training in two steps. We first train without mixup with a learning rate of 5e-5 for three epochs, and then train with mixup starting from previous training's best checkpoint, with a learning rate of 1e-5 for five epochs. 
This two-step training, which also utilized by \citet{zhang2018mixup}, speeds up the model convergence.
We report the best accuracy among both training with and without mixup. For the ANLI task, we select the best checkpoint for training without mixup separately for each round, then conduct training with mixup and report the best accuracy of each round's evaluation dataset.

For each iteration, we split the batch into two smaller batches with the same size, $\batchone$ and $\batchtwo$. Since mixup operation in \ours{} is not symmetric, we conduct mixup back-and-forth so that mixup performance is evaluated regardless of the data position in batch. To prevent the training data distribution getting too far from the original data distribution, we train with and without mixup together as \citet{He2019data}. As a result, we forward each step with average loss from $\batchone$, $\batchtwo$, $mixup(\batchone$, $\batchtwo)$, and $mixup(\batchtwo$, $\batchone)$.

We leave out tokens specific to transformer architecture (e.g., $[CLS]$, $[SEP]$) when conducting a mixup to preserve special signs. As stated by \citet{zhang2018mixup}, giving too high values for mixup ratio may lead to underfitting, while giving $\lambda$ close to 0 leads to the same effect of giving non-augmented original data. From our experiments, we found out that augmentation with prior ratio $\lambda_{0} = 0.1$ is the optimal hyperparameter.

In terms of computation time, \ours{} takes about twice the training time compared with other mixup methods since we need an additional forward and backward step to compute the saliency of tokens. Among hidden-level mixup methods, \tmix{} takes a slightly longer time to train than \emix{}.

\section{Results and Discussion}
\begin{table*}[t!]
\normalsize
\centering
\begin{threeparttable}
\resizebox{\textwidth}{!}{
\begin{tabular}{lccccccccccc} 
    \toprule
    \multicolumn{1}{c}{\multirow{2}{*}{Model}} & 
    \multicolumn{6}{c}{GLUE} & \multicolumn{2}{c}{TREC} & \multicolumn{3}{c}{ANLI} \\
    
    \cmidrule(r{0.125em}){2-7}
    \cmidrule(lr{0.125em}){8-9}
    \cmidrule(l{0.125em}){10-12}
    
     & SST-2 & QQP & MNLI & QNLI & RTE & MRPC & coarse & fine & R1 & R2 & R3 \\
    \midrule
    \multirow{2}{*}{No mixup} & 
    \multirow{2}{*}{92.96} & \multirow{2}{*}{91.32} &
    \multirow{2}{*}{84.27} &  \multirow{2}{*}{91.28} & 
    \multirow{2}{*}{65.56} &     \multirow{2}{*}{86.37} & 
    \multirow{2}{*}{97.08} & \multirow{2}{*}{86.68} & 
    56.40 & 47.10 & 47.62 \\
    & & & & & & & & & 
    57.16 & 47.36 & 48.00 \\ 
    \toprule
    \addlinespace[0.5ex] 
    \multirow{2}{*}{\emix{}} & 
    \multirow{2}{*}{93.03} &     \multirow{2}{*}{91.36} &  
    \multirow{2}{*}{84.35} &     \multirow{2}{*}{91.43} & 
    \multirow{2}{*}{\textbf{67.73}} & 
    \multirow{2}{*}{\textbf{86.72}} & 
    \multirow{2}{*}{97.44} & \multirow{2}{*}{90.04} & 
    56.78 & 47.84 & 47.67 \\
    & & & & & & & & & 
    57.16 & 47.42 & 48.00 \\ 
    %\hline
    \addlinespace[0.5ex]
    \multirow{2}{*}{\tmix{}} & 
    \multirow{2}{*}{93.03} & \multirow{2}{*}{91.34} &  \multirow{2}{*}{84.33} & 
    \multirow{2}{*}{91.40} & \multirow{2}{*}{66.86} & 
    \multirow{2}{*}{86.42} & 
    \multirow{2}{*}{97.52} & \multirow{2}{*}{90.16} & 
    56.68 & 47.58 & 47.78 \\
    & & & & & & & & & 
    57.28 & 47.90 & \textbf{48.42} \\ 
    \toprule
    \addlinespace[0.5ex]
    \multirow{2}{*}{\ours{}} & 
    \multirow{2}{*}{93.10} & 
    \multirow{2}{*}{\textbf{91.43}} & 
    \multirow{2}{*}{\textbf{84.54}} & 
    \multirow{2}{*}{\textbf{91.54}} & 
    \multirow{2}{*}{67.22} & \multirow{2}{*}{86.57} & 
    \multirow{2}{*}{\textbf{97.60}} & 
    \multirow{2}{*}{\textbf{90.24}} & 
    \textbf{57.26} & \textbf{48.36} & 47.78 \\
    & & & & & & & & & 
    \textbf{57.34} & \textbf{48.06} & 48.00 \\
    \addlinespace[0.5ex] 
    \multirow{2}{*}{\ours{} - saliency}&
    \multirow{2}{*}{93.12} &
    \multirow{2}{*}{91.32} &  
    \multirow{2}{*}{84.48} &    
    \multirow{2}{*}{91.29} & 
    \multirow{2}{*}{67.00} & 
    \multirow{2}{*}{86.42} & 
    \multirow{2}{*}{97.44} &
    \multirow{2}{*}{89.56} & 
    57.04 & 48.22 & \textbf{47.95} \\
    & & & & & & & & & 
    57.16 & 47.94 & 48.07 \\ 
    \addlinespace[0.5ex] 
    \multirow{2}{*}{\ours{} - saliency - span}& 
    \multirow{2}{*}{\textbf{93.14}} &
    \multirow{2}{*}{91.32} &  
    \multirow{2}{*}{84.54} &    
    \multirow{2}{*}{91.45} & 
    \multirow{2}{*}{66.93} & 
    \multirow{2}{*}{86.37} & 
    \multirow{2}{*}{97.40} &
    \multirow{2}{*}{89.20} & 
    56.74 & 47.52 & 47.77 \\
    & & & & & & & & & 
    57.20 & 47.90 & 48.00 \\ 
    \bottomrule
\end{tabular}}
\end{threeparttable}
\caption{
Experimental results of comparison with baselines and ablation study. 
All values are average accuracy (\%) of five runs with different seeds. 
MNLI indicates MNLI-mismatched dev set accuracy. 
We report validation accuracy for GLUE, test accuracy for TREC, and valid (upper) / test (lower) accuracy for ANLI. We report variance on Appendix.~\ref{section:variance}.
}
\label{tab:result}
\vspace{-5pt}

\end{table*}

\begin{figure}[!t]
    \centering
    \begin{adjustbox}{max size={0.45\textwidth}{0.45\textheight}}
    \includegraphics{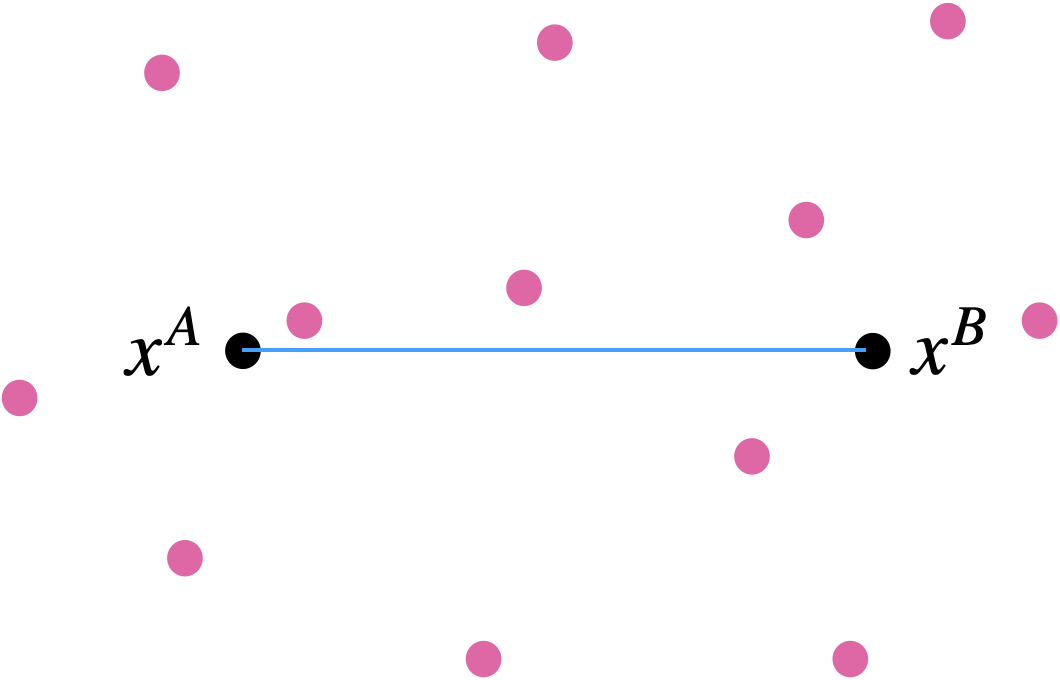}
    \end{adjustbox}
    \caption{
    Visualization of original data and synthesized data by hidden-level mixup (EmbedMix or TMix) and \ours{} in the hidden space. Black dots indicate the original data, $\xzero{}$ and $\xone{}$. For hidden-level mixup, synthetic data ($\tilde{x}$) are created only along the line (blue) connecting two points, since it is a linear combination within the hidden space. However, \ours{} explore larger synthetic sample space for $\tilde{x}$, since it consists of a discrete combination within the \textit{input} space. Synthetic data for \ours{} are illustrated in pink dots.
    }
    \label{fig:data_space}
    \vspace{-5pt}
\end{figure}

Table~\ref{tab:result} illustrates our results. We investigate the effectiveness of \ours{} compared with hidden layer mixup methods on the aspect of dataset size, number of class labels, and paired sentence tasks. 
\paragraph{Dataset size}
Compared with hidden-level mixup methods, \ours{} fully demonstrate its effectiveness on datasets having a sufficient amount of data. Since \ours{} is a discrete combination rather than a linear combination of two data samples, it creates data samples on a synthetic space in a larger range than hidden-level mixup (Fig.~\ref{fig:data_space}). We hypothesize that a large amount of data help better representation in synthetic space. 

\paragraph{The number of class labels} 
\ours{} is especially effective for multiple class label datasets (TREC, ANLI, MNLI, QNLI). Accordingly, the accuracy gain of \ours{} from the training without mixup is much higher on TREC-fine (47 labels) than TREC-coarse (6 labels), with +3.56 and +0.52, respectively.
We hypothesize that this result originates from the mixup characteristic that benefits more from cross-label mixup than mixup with the same label, as stated at \citet{zhang2018mixup}.\footnote{ \citet{zhang2018mixup} states that mixing random pairs from all classes (per-batch basis) has the strongest regularization effect compared with mixup by per-class (same class) basis. } 
Since datasets with multiple total class labels increase the possibility of being selected cross-label in a random sampling of mixup sources, we assert mixup performance increases in such datasets.
\paragraph{Paired sentence tasks}
\ours{} have a competitive advantage on paired sentence tasks, such as textual entailment or similarity classification.
We suspect this accuracy gain originates from consideration of individual tokens.
Existing methods (hidden-level mixup) apply mixup on the hidden layer, without consideration of special tokens, i.e., $[SEP], [CLS]$. These methods may lose information about the start of the sentence or appropriate separation of pair of sentences. 
In contrast, \ours{} can consider the individual token property when applying mixup. Here, our mixup strategy on paired data (Section~\ref{section:salmix}) preserves the property of $[SEP]$, which is not guaranteed by hidden mixup. 

\paragraph{Ablation Study}
The results of \ours{} and its variants demonstrate that the performance improves as we add span constraint and saliency information. Adding span constraint in the mixup operation benefit from better localizable ability, and most salient spans have more relationship to corresponding labels while discarding least salient spans have a higher probability that those spans are not semantically important with respect to the original labels. 
Among those two, introducing saliency information contributes to accuracy relatively more than the span constraint.

\section{Conclusion}
We present \ours{}, a novel and simple input-level mixup method for text data that improves regularization ability leading to better performance in text classification. 
\ours{} preserves the locality of mixing texts by replacing in span-level and keep most discriminative tokens in the mixed text using saliency score.
Throughout the experiment, we show that our method improves performance in various types of text classification tasks. 
For future work, we plan to apply \ours{} on a broader range of tasks, including generation or different scenarios like semi-supervised learning. 

\section*{Acknowledgments}
The authors would like to thank Clova AI members for proofreading this manuscript and the anonymous reviewers for their constructive feedback.
We use Naver Smart Machine Learning \cite{sung2017nsml,kim2018nsml} platform for the experiments. This work was supported by Institute of Information \& communications Technology Planning \& Evaluation (IITP) grant funded by the Korea government (MSIT) 
(No.2019-0-00075, Artificial Intelligence Graduate School Program (KAIST)).
\noindent

\clearpage

\bibliographystyle{acl_natbib}
\bibliography{anthology,acl2021}

\clearpage
\appendix

\numberwithin{table}{section}
\setcounter{page}{1}
\onecolumn

\section{Accuracy Variance}
\label{section:variance}
\begin{table*}[h]
\centering
\begin{threeparttable}
\resizebox{\textwidth}{!}{
\begin{tabular}{lccccccccccc} 
    \toprule
    \multicolumn{1}{c}{\multirow{2}{*}{Model}} & 
    \multicolumn{6}{c}{GLUE} & \multicolumn{2}{c}{TREC} & \multicolumn{3}{c}{ANLI} \\
    
    \cmidrule(r{0.125em}){2-7}
    \cmidrule(lr{0.125em}){8-9}
    \cmidrule(l{0.125em}){10-12}
    
     & SST-2 & QQP & MNLI & QNLI & RTE & MRPC &  Coarse & Fine & R1 & R2 & R3 \\
    \midrule
    \multirow{2}{*}{No mixup} & 
    \multirow{2}{*}{0.04} & \multirow{2}{*}{0.04} &
    \multirow{2}{*}{0.12} &  \multirow{2}{*}{0.05} & 
    \multirow{2}{*}{3.89} &     \multirow{2}{*}{1.73} & 
    \multirow{2}{*}{0.17} & \multirow{2}{*}{2.21} & 
    1.21 & 0.16 & 0.73 \\
    & & & & & & & & & 
    0.24 & 1.26 & 0.84 \\ 
    % \hline
    \toprule
    %\addlinespace[0.5ex] 
    \multirow{2}{*}{\emix{}} & 
    \multirow{2}{*}{0.02} &     \multirow{2}{*}{0.03} &  
    \multirow{2}{*}{0.14} &     \multirow{2}{*}{0.04} & 
    \multirow{2}{*}{3.89} & 
    \multirow{2}{*}{1.39} & 
    \multirow{2}{*}{0.09} & \multirow{2}{*}{0.31} & 
    1.38 & 0.46 & 0.75 \\
    & & & & & & & & & 
    0.24 & 1.18 & 0.84 \\ 
    % \hline
    %\addlinespace[0.5ex]
    \multirow{2}{*}{\tmix{}} & 
    \multirow{2}{*}{0.04} & \multirow{2}{*}{0.04} &  \multirow{2}{*}{0.09} & 
    \multirow{2}{*}{0.03} & \multirow{2}{*}{1.85} & 
    \multirow{2}{*}{1.55} & 
    \multirow{2}{*}{0.05} & \multirow{2}{*}{0.63} & 
    1.44 & 0.33 & 0.73 \\
    & & & & & & & & & 
    0.25 & 0.75 & 1.28 \\ 
    % \hline
    % \hline
    \toprule
    %\addlinespace[0.5ex]
    \multirow{2}{*}{\ours{}} & 
    \multirow{2}{*}{0.03} & 
    \multirow{2}{*}{0.07} & 
    \multirow{2}{*}{0.07} & 
    \multirow{2}{*}{0.03} & 
    \multirow{2}{*}{2.57} & 
    \multirow{2}{*}{1.15} & 
    \multirow{2}{*}{0.03} & 
    \multirow{2}{*}{0.49} & 
    1.56& 0.27 & 0.73 \\
    & & & & & & & & & 
    0.25 & 0.46 & 0.84 \\
    % \hline 
    %\addlinespace[0.5ex]
    \multirow{2}{*}{\ours{} - saliency} & 
    \multirow{2}{*}{0.02} & 
    \multirow{2}{*}{0.04} & 
    \multirow{2}{*}{0.11} & 
    \multirow{2}{*}{0.04} & 
    \multirow{2}{*}{2.06} & 
    \multirow{2}{*}{1.55} & 
    \multirow{2}{*}{0.09} & 
    \multirow{2}{*}{0.69} & 
    1.33 & 0.18 & 0.62 \\
    & & & & & & & & & 
    0.24 & 1.99 & 0.80 \\
    % \hline
    %\addlinespace[0.5ex]
    \multirow{2}{*}{\ours{} - saliency - span} & 
    \multirow{2}{*}{0.00} & 
    \multirow{2}{*}{0.04} & 
    \multirow{2}{*}{0.09} & 
    \multirow{2}{*}{0.03} & 
    \multirow{2}{*}{1.86} & 
    \multirow{2}{*}{1.73} & 
    \multirow{2}{*}{0.08} & 
    \multirow{2}{*}{0.14} & 
    2.01 & 0.11 & 0.68 \\
    & & & & & & & & & 
    0.28 & 0.45 & 0.84 \\
    \bottomrule
\end{tabular}}
\end{threeparttable}
\caption{Standard deviation results, corresponding with the average of our experiments. The deviation is conducted by 5 runs with different seeds.}
\label{tab:variance}
\end{table*}

\noindent  We also report accuracy variance among the five seeds for each experiment (Table.~\ref{tab:variance}).

\section{Ablation}
\label{section:ablation_general}
Fig.~\ref{fig:ablation_overview} and Fig.~\ref{fig:ablation_unk} shows the illustration of different variants of \ours{} and random UNK replacement with $\lambda = 0.2$. Fig.~\ref{fig:paired_sentence_example} shows the illustration of getting the augmented output with lambda calculation by \ours{} for paired sentence tasks.
The saliency maps are visualized where darker concentration of colors mean higher contribution to corresponding label. 

\subsection{Variants of \ours{}}
\label{section:ablation}
\begin{figure}[!h]
\centering
\subfloat[Normal training without mixup]{\label{c}\includegraphics[width=.45\linewidth]{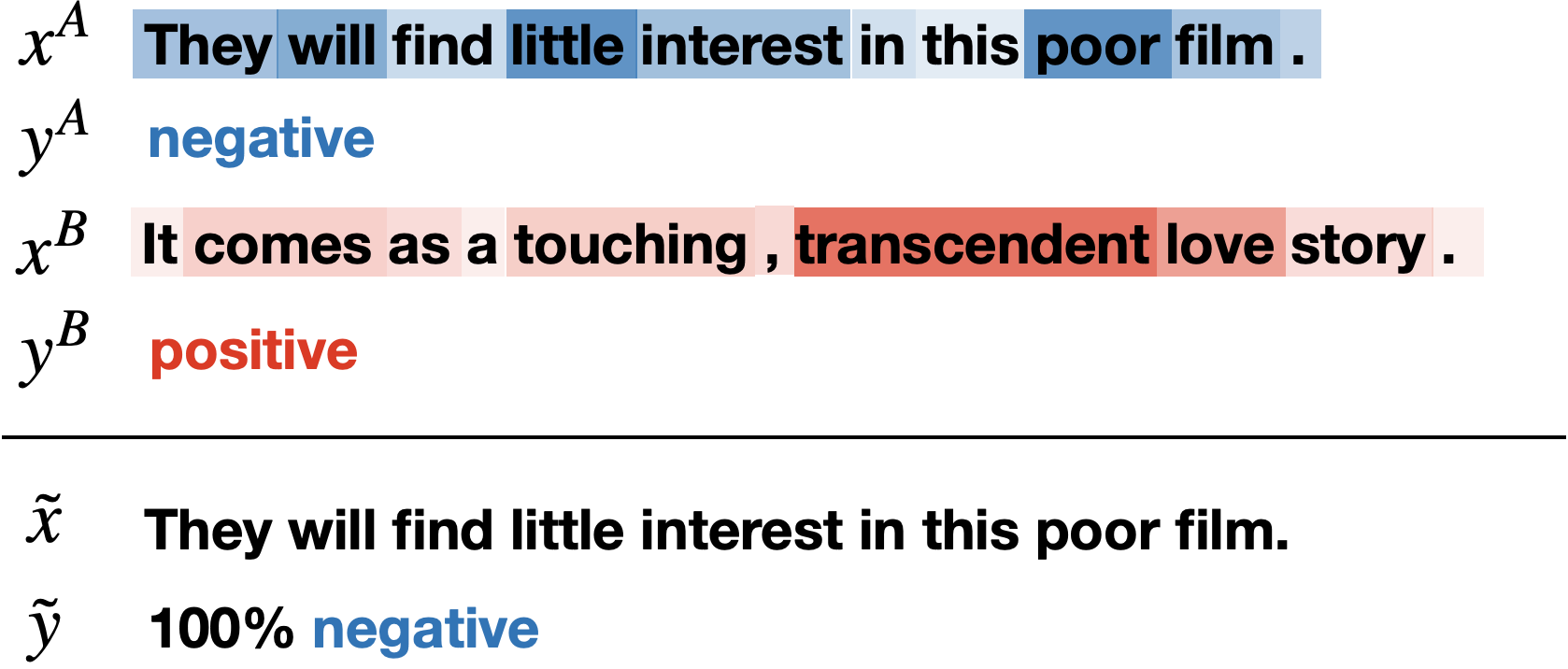}}\par 
\vspace{0.5cm}
\subfloat[\ours{} - saliency]{\label{b}\includegraphics[width=.45\linewidth]{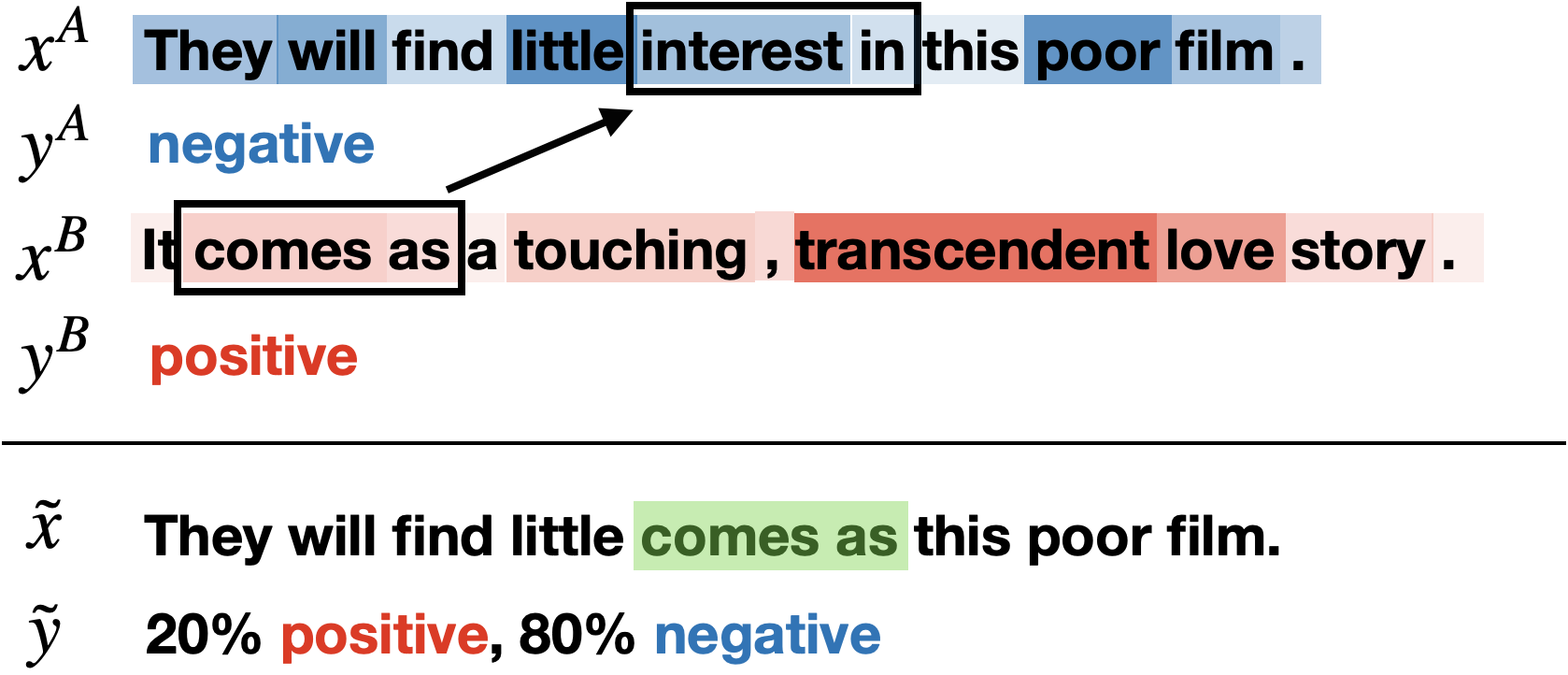}}
\hspace{5mm}
\subfloat[\ours{} - saliency - span]{\label{a}\includegraphics[width=.45\linewidth]{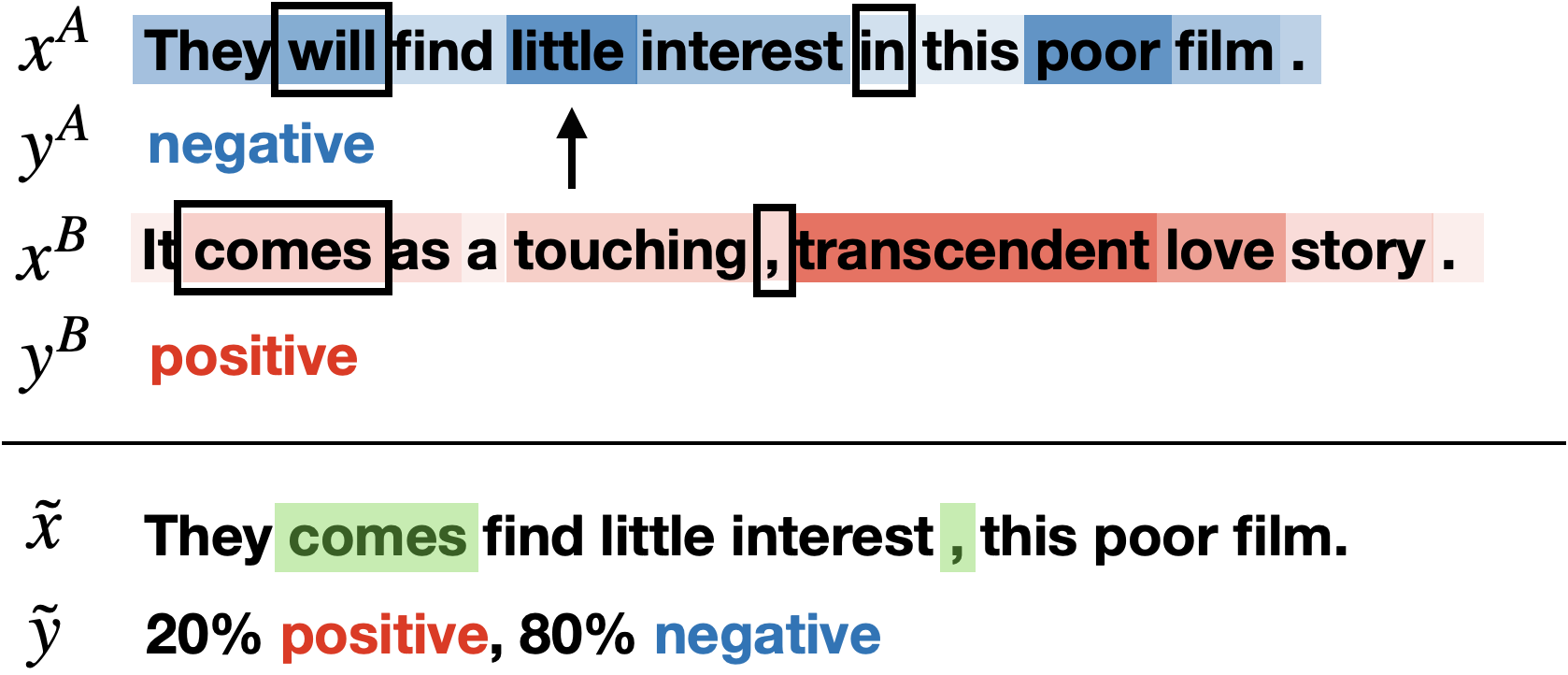}}\hfill
\caption{Illustration of normal training and variants of \ours{}}
\label{fig:ablation_overview}
\end{figure}

Here, we describe in detail how we implement \ours{} without saliency (Figure.~\ref{fig:ablation_overview} (b)) and \ours{} without saliency and span restriction (Figure.~\ref{fig:ablation_overview} (c)).

At normal training, only two real data samples ($\xzero{}$ and $\xone{}$) are used to train the model. For Figure.~\ref{fig:ablation_overview} (b), we \textit{randomly} select each span from $\xzero{}$ and $\xone{}$. Then, we replace $\xone{}$ to $\xzero{}$ to make a new data $\tilde{x}$.
For Figure.~\ref{fig:ablation_overview} (c), input level mixup is conducted on a per-token basis. After calculaton of $l$ given the prior mixup ratio, we randomly sample tokens from $\xzero{}$. The tokens need not be a contiguous span. Then, we replace tokens accordingly with the position of the token be preserved, meaning that the second token from $\xzero{}$ is replaced with second token from $\xone{}$, the sixth token from $\xzero{}$ is replaced with sixth token from $\xone{}$ (by the illustration example), and so on.

\subsection{Comparison with other simple augmentation methods}
\label{section:ablation_unk}
\begin{figure}[!h]
\centering
\subfloat[Random \texttt{[UNK]} replacement]{\label{a}\includegraphics[width=.45\linewidth]{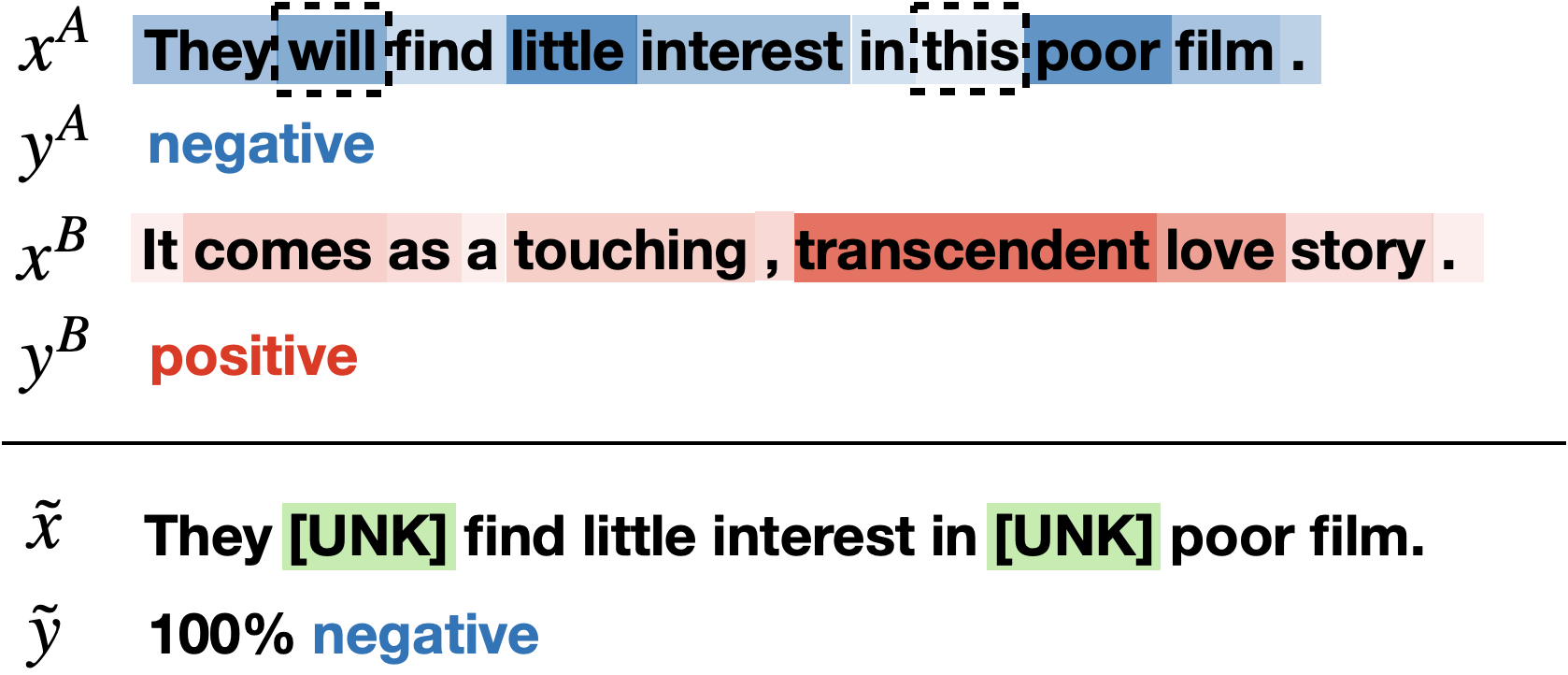}}
\hspace{5mm}
\subfloat[\ours{}]{\label{b}\includegraphics[width=.45\linewidth]{images/new-SalMix.png}}\hfill

\caption{Comparison of our methods with word dropout}
\label{fig:ablation_unk}
\end{figure}
We also compare \ours{} with simple word dropout methods, which may seem similar in the perspective that they create noisy sentences. The difference is whether label mixup is performed. Illustration of the implementation of random [UNK] replacement is available at Fig.~\ref{fig:ablation_unk}. 
Random UNK replacement is similar to word dropout. We don't use $\xone{}$ when making synthetic samples ($l$ = 0). Instead, we randomly sample a set of tokens from $\xzero{}$ and replace each token in that span with [UNK]. The process is similar to Figure.~\ref{fig:ablation_overview} (c), except that the selected tokens at $\xzero{}$ are replaced into [UNK]. Another difference is that the output label ($\tilde{y}$) completely follow the origin ($y^A$) and no label mixup is performed. The illustration is available at ~\ref{fig:ablation_overview}.

We evaluate the random [UNK] replacement method on all dataset with \ours{} and variants of \ours{} at ablation study. By the experiment results at Table ~\ref{tab:ablation}, we show that input level mixup methods generally outperform simple regularization methods. This means that datasets synthesized from \ours{} and the according target vectors have more gain on the generalization ability than word dropout.
\begin{table*}[t!]
\normalsize
\centering
\begin{threeparttable}
\resizebox{\textwidth}{!}{
\begin{tabular}{lccccccccccc} 
    \toprule
    \multicolumn{1}{c}{\multirow{2}{*}{Model}} & 
    \multicolumn{6}{c}{GLUE} & \multicolumn{2}{c}{TREC} & \multicolumn{3}{c}{ANLI} \\
    
    \cmidrule(r{0.125em}){2-7}
    \cmidrule(lr{0.125em}){8-9}
    \cmidrule(l{0.125em}){10-12}
    
     & SST-2 & QQP & MNLI & QNLI & RTE & MRPC & coarse & fine & R1 & R2 & R3 \\
    \midrule
    \multirow{2}{*}{No mixup} & 
    \multirow{2}{*}{92.96} & \multirow{2}{*}{91.32} &
    \multirow{2}{*}{84.27} &  \multirow{2}{*}{91.28} & 
    \multirow{2}{*}{65.56} &     \multirow{2}{*}{86.37} & 
    \multirow{2}{*}{97.08} & \multirow{2}{*}{86.68} & 
    56.40 & 47.10 & 47.62 \\
    & & & & & & & & & 
    57.16 & 47.36 & 48.00 \\  
    \toprule
    \addlinespace[0.5ex]
    \multirow{2}{*}{Random UNK replacement} & 
    \multirow{2}{*}{93.10} & \multirow{2}{*}{91.33} &  \multirow{2}{*}{84.46} & 
    \multirow{2}{*}{91.45} & \multirow{2}{*}{66.86} & 
    \multirow{2}{*}{\textbf{86.62}} & 
    \multirow{2}{*}{97.44} & \multirow{2}{*}{89.24} & 
    56.98 & 47.86 & \textbf{47.98} \\
    & & & & & & & & & 
    57.26 & \textbf{48.36} & \textbf{48.32} \\ 
    \toprule
    \addlinespace[0.5ex]
    \multirow{2}{*}{\ours{}} & 
    \multirow{2}{*}{93.10} & 
    \multirow{2}{*}{\textbf{91.43}} & 
    \multirow{2}{*}{\textbf{84.54}} & 
    \multirow{2}{*}{\textbf{91.54}} & 
    \multirow{2}{*}{\textbf{67.22}} & \multirow{2}{*}{86.57} & 
    \multirow{2}{*}{\textbf{97.60}} & 
    \multirow{2}{*}{\textbf{90.24}} & 
    \textbf{57.26} & \textbf{48.36} & 47.78 \\
    & & & & & & & & & 
    \textbf{57.34} & 48.06 & 48.00 \\
    \addlinespace[0.5ex] 
    \multirow{2}{*}{\ours{} - saliency}&
    \multirow{2}{*}{93.12} &
    \multirow{2}{*}{91.32} &  
    \multirow{2}{*}{84.48} &    
    \multirow{2}{*}{91.29} & 
    \multirow{2}{*}{67.00} & 
    \multirow{2}{*}{86.42} & 
    \multirow{2}{*}{97.44} &
    \multirow{2}{*}{89.56} & 
    57.04 & 48.22 & 47.95 \\
    & & & & & & & & & 
    57.16 & 47.94 & 48.07 \\ 
    \addlinespace[0.5ex] 
    \multirow{2}{*}{\ours{} - saliency - span}& 
    \multirow{2}{*}{\textbf{93.14}} &
    \multirow{2}{*}{91.32} &  
    \multirow{2}{*}{84.54} &    
    \multirow{2}{*}{91.45} & 
    \multirow{2}{*}{66.93} & 
    \multirow{2}{*}{86.37} & 
    \multirow{2}{*}{97.40} &
    \multirow{2}{*}{89.20} & 
    56.74 & 47.52 & 47.77 \\
    & & & & & & & & & 
    57.20 & 47.90 & 48.00 \\ 
    \bottomrule
\end{tabular}}
\end{threeparttable}
\caption{Accuracy (\%) comparison with simple data augmentation method(random UNK replacement) and input mixup methods. The results are average of five runs with different seeds. Results show that our input level mixup methods are generally competitive with simple word dropout methods.}
\label{tab:ablation}

\end{table*}

\subsection{Illustration of SSMix on paired sentence tasks}
\label{appendix:paired_example}
\begin{figure}[!t]
    \centering
    \begin{adjustbox}{max size={0.5\textwidth}{0.5\textheight}}
    \includegraphics{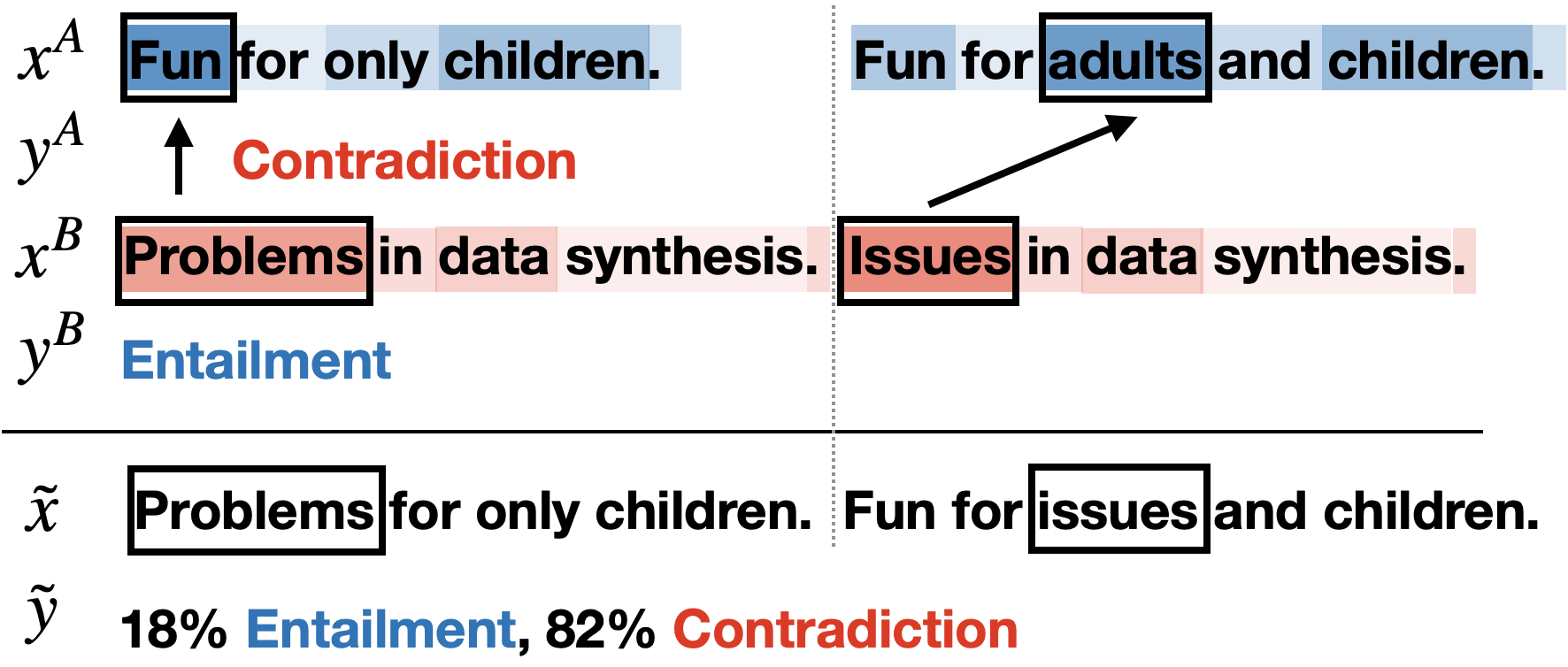}
    \end{adjustbox}
    \caption{
    Illustration of applying SSMix to make $\tilde{x}$ for paired sentence, in particular NLI tasks, which classifies whether the relation of sentence pairs is entailment, neutral, or contradiction. Mixup is conducted individually, sentence by sentence.}
    \label{fig:paired_sentence_example}
    \vspace{-5pt}
\end{figure}
Fig. \ref{fig:paired_sentence_example} shows the illustration of example for paired sentence. Here, "Fun for only children." and "Fun for adults and children." correspond to $p^{A}$ and $q^{A}$, "Problems in data synthesis." and "Issues in data synthesis." correspond to $p^{B}$ and $q^{B}$, and "Problems for only children.", "Fun for issues and children." correspond to $p$ and $q$, respectively. $\lambda$ is calculated as : $\lambda = (|p_{S}| + |q_{S}|) / (|\tilde{p}| + |\tilde{q}|) = (1 + 1)/(5 + 6) = 2 / 11 \approx 0.18$.

\end{document}